\documentclass{article}

\usepackage{PRIMEarxiv}

\usepackage[utf8]{inputenc} 
\usepackage[T1]{fontenc}    
\usepackage{hyperref}       
\usepackage{url}            
\usepackage{booktabs}       
\usepackage{amsfonts}       
\usepackage{nicefrac}       
\usepackage{microtype}      
\usepackage{lipsum}
\usepackage{fancyhdr}       
\usepackage{graphicx}       
\graphicspath{{media/}}     

\usepackage{subcaption}

\usepackage{bm} 
\usepackage{bbm}
\usepackage{amsmath}

\title{Cross-Model Transfer of Task Vectors via Few-Shot Orthogonal Alignment
}

\author{
  Kazuhiko Kawamoto,\quad Atsuhiro Endo\thanks{Currently at Hitachi, Ltd.},\quad Hiroshi Kera\\
  Chiba University \\
  \texttt{kawa@faculty.chiba-u.jp, \{a.endo, kera\}@chiba-u.jp} \\
}

\begin{document}
\maketitle

\begin{abstract}
Task arithmetic enables efficient model editing by representing task-specific changes as vectors in parameter space. Task arithmetic typically assumes that the source and target models are initialized from the same pre-trained parameters. This assumption limits its applicability in cross-model transfer settings, where models are independently pre-trained on different datasets. To address this challenge, we propose a method based on few-shot orthogonal alignment, which aligns task vectors to the parameter space of a differently pre-trained target model. These transformations preserve key properties of task vectors, such as norm and rank, and are learned using only a small number of labeled examples. We evaluate the method using two Vision Transformers pre-trained on YFCC100M and LAION400M, and test on eight classification datasets. Experimental results show that our method improves transfer accuracy over direct task vector application and achieves performance comparable to few-shot fine-tuning, while maintaining the modularity and reusability of task vectors. Our code is available at \url{https://github.com/kawakera-lab/CrossModelTransfer}.
\end{abstract}


\section{Introduction}
Pre-trained models provide general-purpose representations learned from large datasets~\cite{kornblith2019better, radford2021learning, caron2021emerging, kirillov2023segment}, and are widely used in downstream tasks.
Fine-tuning improves task-specific performance but requires additional storage and computation, since each task requires maintaining a separate fine-tuned model.

To reduce these costs, task arithmetic~\cite{ilharco2023editing} defines task vectors as the difference between the parameters of a fine-tuned model and its original pre-trained model. Adding or subtracting task vectors enables models to acquire or forget task-specific capabilities. 
Task arithmetic builds on prior insights into the structure of parameter space. For example, low-loss interpolation~\cite{wortsman2022model} shows that averaging parameters from similarly initialized models can yield good performance, and mode connectivity~\cite{frankle2020linear} demonstrates that two independently trained models can be connected by a nonlinear path in parameter space that maintains low loss. These findings suggest that tasks can be interpreted as directions or trajectories in parameter space~\cite{gupta2022model}, which task arithmetic exploits by modeling task-specific changes as vectors.

However, task arithmetic assumes that all models share the same pre-trained base. When model initialization or training data differs, performance degrades significantly~\cite{ainsworth2023git} because parameter spaces no longer align and task vectors fail to transfer. 
To address this, recent works have proposed model alignment techniques. Git Re-Basin~\cite{ainsworth2023git}, for example, aligns models via neuron permutations.
ZipIt!~\cite{stoica2024zipit} proposes a model merging technique that combines models trained on different tasks without additional training. 
While these methods demonstrate the feasibility of model merging, they primarily focus on full-model integration rather than the reuse of task vectors.

In this work, we tackle the challenge of cross-model task vector transfer by introducing 
orthogonal similarity transformations that explicitly align parameter subspaces before applying task updates. Using only a few labeled examples, we estimate layer-wise orthogonal matrices that rotate task vectors into the target model’s parameter space, while preserving their norm and rank. 

We evaluate the proposed method on a benchmark of eight classification datasets using two Vision Transformers pre-trained on different datasets. 
The experimental results show improved transfer accuracy compared to direct task vector application. 
The method achieves accuracy comparable to few-shot fine-tuning, while preserving the modularity and reusability of task vectors.

\section{Related Work}

Task arithmetic \cite{ilharco2023editing} enables neural network editing by representing task-specific changes as vectors in parameter space. A task vector is defined as the difference between a fine-tuned model and its pre-trained counterpart. Adding or subtracting these vectors allows a model to acquire or remove task-specific capabilities. This approach supports efficient multi-task learning and selective forgetting.

Several methods aim to enhance task vector usage and model editing techniques.  
Ortiz-Jimenez et al.~\cite{ortizjimenez2023tangent} improve robustness by operating in the tangent space of the pre-trained model.  
Iurada et al.~\cite{iurada2025sparse} propose a sparse fine-tuning strategy that updates only a subset of parameters, improving efficiency and modularity.  
ADAMerging~\cite{yang2024adamerging} adaptively combines task-specific updates using importance parameters to reduce task interference.  
Localize-and-Stitch~\cite{he2025localizeandstitch} leverages local geometric structure to support modular task composition.  
While these studies offer valuable insights, most do not explicitly address the challenge of transferring task vectors across models with different pre-training.

Other work supports task arithmetic by examining the geometry of neural network parameters.  
Wortsman et al.~\cite{wortsman2022model} demonstrate that averaging parameters from similarly initialized models maintains performance.  
Frankle et al.~\cite{frankle2020linear} introduce mode connectivity, showing that low-loss paths can exist between trained models.  
Gupta et al.~\cite{gupta2022model} further argue that parameter updates can be interpreted as geometric directions in parameter space.

However, task arithmetic typically assumes that both the source and target models are initialized from the same pre-trained parameters. When pre-training differs, direct task vector transfer often fails due to misaligned parameter spaces.  
Several methods have been proposed to address this issue.  
Git Re-Basin~\cite{ainsworth2023git} aligns neurons through permutation.  
Matena and Raffel~\cite{matena2022merging} introduce Fisher-weighted averaging for model merging.  
Foldable SuperNets~\cite{kinderman2024foldable} optimize a shared SuperNet to merge transformers with different initializations.  
ZipIt!~\cite{stoica2024zipit} proposes a feature-space “zipping” operation to enable partial parameter sharing while preserving task-specific functionality.  
These methods, however, focus on merging full models rather than reusing task vectors.

Recent studies have explored how to improve task vector composition and generalization.  
Zhang et al.~\cite{zhang2024atlas} propose anisotropic scaling to merge task vectors from different tasks or domains.  
Luo et al.~\cite{luo2025crossmodal} demonstrate the feasibility of transferring task vectors across modalities, such as vision and language.  
Sun et al.~\cite{sun2025trustregion} introduce a trust-region-based approach to mitigate interference during task vector application.  
Li et al.~\cite{li2025provable} identify theoretical conditions under which task vectors remain effective, particularly in non-linear models.

Our work directly addresses the challenge of transferring task vectors across pre-trained model boundaries. Instead of merging entire models, we align and reuse task vectors between differently pre-trained models.  
We propose a lightweight few-shot method based on orthogonal transformations that enables effective adaptation of task vectors using minimal supervision.

\section{Method}

We propose a method for transferring task vectors between models with different pre-training. While task vectors are computed using standard fine-tuning techniques on a source model, our contribution lies in aligning these vectors to the target model’s parameter space using orthogonal similarity transformations. This alignment preserves important properties such as norm and rank, and enables accurate transfer without modifying the target model's architecture.

\subsection{Task Vector Computation via Selective Fine-Tuning}

We construct task vectors by fine-tuning selected parameter matrices of a pre-trained Vision Transformer (ViT) \cite{dosovitskiy2021an}, keeping all other parameters fixed. We focus on two parameter groups: the attention embedding layers and LoRA \cite{hu2022lora} matrices. This design reduces computation while retaining task-specific adaptation.
We also evaluate task vectors computed from all parameters. As shown in Appendix~\ref{appendix:full weight}, this yields lower transfer performance than using only embedding layers or LoRA modules.

\subsubsection{Embedding-Based Task Vectors}

In the embedding setting, we fine-tune only the attention embedding layers in the self-attention blocks of the ViT. Each target parameter matrix $\mathbf{W}$ corresponds to one of the projection matrices in the attention mechanism. All other parameters remain frozen.

For each such matrix, the task vector is defined as:
\begin{equation}
\Delta \mathbf{W} = \mathbf{W}' - \mathbf{W},
\end{equation}
where $\mathbf{W}'$ is the fine-tuned parameter. This update is applied to each self-attention block, and the task vector is defined as the collection of resulting layer-wise differences.

\subsubsection{LoRA-Based Task Vectors}

In the LoRA setting, we insert trainable low-rank matrices into the same attention embedding layers and fine-tune only these added parameters. For each matrix $\mathbf{W}$, LoRA introduces low-rank matrices $\mathbf{A} \in \mathbb{R}^{d \times r}$ and $\mathbf{B} \in \mathbb{R}^{r \times d}$ such that:
\begin{equation}
\mathbf{W}' = \mathbf{W} + \mathbf{A} \mathbf{B}.
\end{equation}
The corresponding task vector is:
\begin{equation}
\Delta \mathbf{W} = \mathbf{W}' - \mathbf{W} = \mathbf{A} \mathbf{B}.
\end{equation}
These low-rank updates are computed for each relevant attention matrix in each self-attention block, and together they constitute the full task vector.

\subsection{Orthogonal Similarity Transformation for Cross-Model Alignment}

We apply layer-wise orthogonal similarity transforms to align task updates with the target model without changing their norm or rank.
Let \(\Delta \mathbf{W}^{(S)}_l\) be the task vector for layer \(l\) of the source model (\(l=1,\dots,L\)), and let \(\mathbf{U}_l\in\mathbb{O}(d_l)\) be the corresponding orthogonal matrix. The transformed update for layer \(l\) in the target model is
\[
\Delta \mathbf{W}^{(T)}_l
= \mathbf{U}_l^\top\,\Delta \mathbf{W}^{(S)}_l\,\mathbf{U}_l.
\]
where \( \Delta \mathbf{W}^{(T)}_l \) is the aligned task vector in the target model’s parameter space.

We learn all orthogonal matrices \(\{\mathbf{U}_l\}\) by minimizing the total cross-entropy loss and an orthogonality penalty:
\[
\mathcal{L}_{\mathrm{total}}
= \sum_{l=1}^L 
   \mathcal{L}_{\mathrm{CE}}\Bigl(\mathbf{W}^{(T)}_l + \mathbf{U}_l^\top \Delta \mathbf{W}^{(S)}_l\,\mathbf{U}_l\Bigr)
\;+\;\alpha
\sum_{l=1}^L 
   \bigl\|\mathbf{U}_l^\top\mathbf{U}_l - \mathbf{I}\bigr\|_F^2,
\]
where \(\mathbf{I}\) is the identity matrix, \(\|\cdot\|_F\) denotes the Frobenius norm, and \(\alpha\) trades off between the two terms. Each \(\mathbf{U}_l\) is initialized to \(\mathbf{I}\), so early updates leave the task vectors unchanged—an advantage in few-shot regimes.

\subsection{Why Orthogonal Similarity Transformations?}

Orthogonal similarity transformations offer several theoretical advantages by preserving geometric and algebraic properties of a linear mapping.

First, they preserve the Frobenius norm:
\begin{equation}
\|\mathbf{U}^\top \Delta \mathbf{W}\,\mathbf{U}\|_F \;=\;\|\Delta \mathbf{W}\|_F,
\end{equation}
which prevents any unintended scaling of task updates.
Second, they preserve matrix rank:
\begin{equation}
\operatorname{rank}(\mathbf{U}^\top \Delta \mathbf{W}\,\mathbf{U}) \;=\;\operatorname{rank}(\Delta \mathbf{W}),
\end{equation}
ensuring that the low-rank structure of LoRA-based vectors is maintained.
Third, they compensate for rotational mismatches between independently pre-trained models. If the source and target parameter matrices differ by a rotation,
\begin{equation}
\mathbf{W}^{(T)} \approx \mathbf{U}^\top \,\mathbf{W}^{(S)}\,\mathbf{U},
\end{equation}
then a task vector computed in the source model,
\(\Delta \mathbf{W}^{(S)}\), can be aligned to the target model’s coordinate system via
\begin{equation}
\Delta \mathbf{W}^{(T)} \approx \mathbf{U}^\top\,\Delta \mathbf{W}^{(S)}\,\mathbf{U}.
\end{equation}

\section{Experiments}

We evaluate the effectiveness of our orthogonal transformation method for transferring task vectors across models with different pre-training. Our experiments demonstrate that combining orthogonal transformations with few-shot learning enables accurate task transfer even under significant domain shift.

\subsection{Experimental Setup}

\paragraph{Pre-trained Models.}
We use two ViT-B/32 models with different pre-training datasets: YFCC100M~\cite{thomee2016yfcc100m} and LAION400M~\cite{schuhmann2021laion400m}. This setup introduces a domain gap due to differences in visual content distributions. The YFCC100M-pretrained model is used as the source model, and the LAION400M-pretrained model is used as the target. All task vectors are computed by fine-tuning the source model and then applied to the target model for evaluation.

\begin{figure}[t]
    \centering
    \begin{minipage}{0.2\linewidth}
        \centering
        \includegraphics[width=0.9\linewidth]{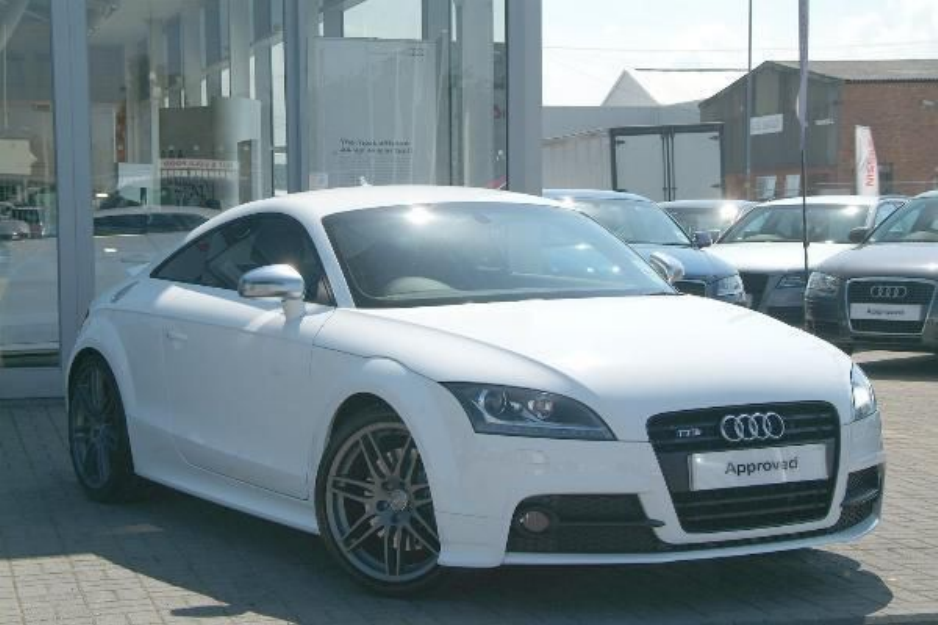}
        \subcaption{Cars}
        \label{fig:cars}
    \end{minipage}
    \begin{minipage}{0.2\linewidth}
        \centering
        \includegraphics[width=0.9\linewidth]{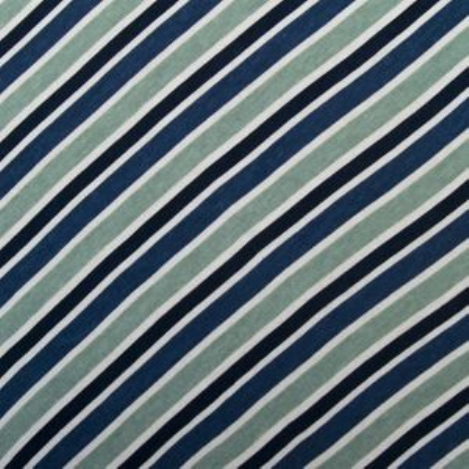}
        \subcaption{DTD}
        \label{fig:dtd}
    \end{minipage}
    \begin{minipage}{0.2\linewidth}
        \centering
        \includegraphics[width=0.9\linewidth]{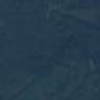}
        \subcaption{EuroSAT}
        \label{fig:eurosat}
    \end{minipage}
    \begin{minipage}{0.2\linewidth}
        \centering
        \includegraphics[width=0.9\linewidth]{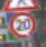}
        \subcaption{GTSRB}
        \label{fig:gtsrb}
    \end{minipage}

    \vspace{1mm}

    \begin{minipage}{0.2\linewidth}
        \centering
        \includegraphics[width=0.9\linewidth]{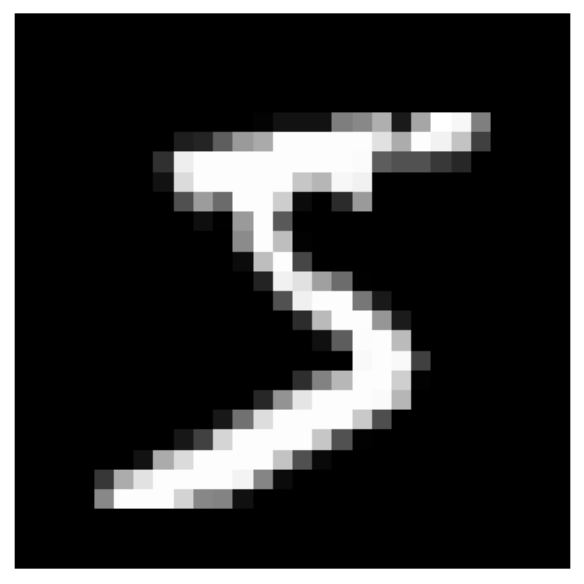}
        \subcaption{MNIST}
        \label{fig:mnist}
    \end{minipage}
    \begin{minipage}{0.2\linewidth}
        \centering
        \includegraphics[width=0.9\linewidth]{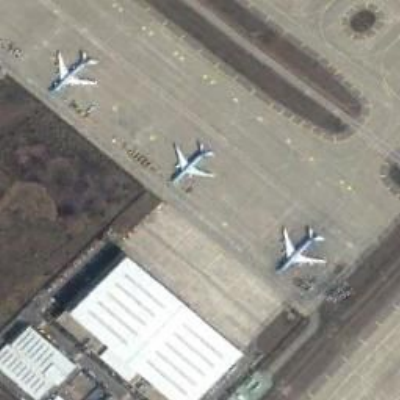}
        \subcaption{RESISC45}
        \label{fig:resisc45}
    \end{minipage}
    \begin{minipage}{0.2\linewidth}
        \centering
        \includegraphics[width=0.9\linewidth]{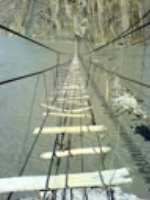}
        \subcaption{SUN397}
        \label{fig:sun397}
    \end{minipage}
    \begin{minipage}{0.2\linewidth}
        \centering
        \includegraphics[width=0.9\linewidth]{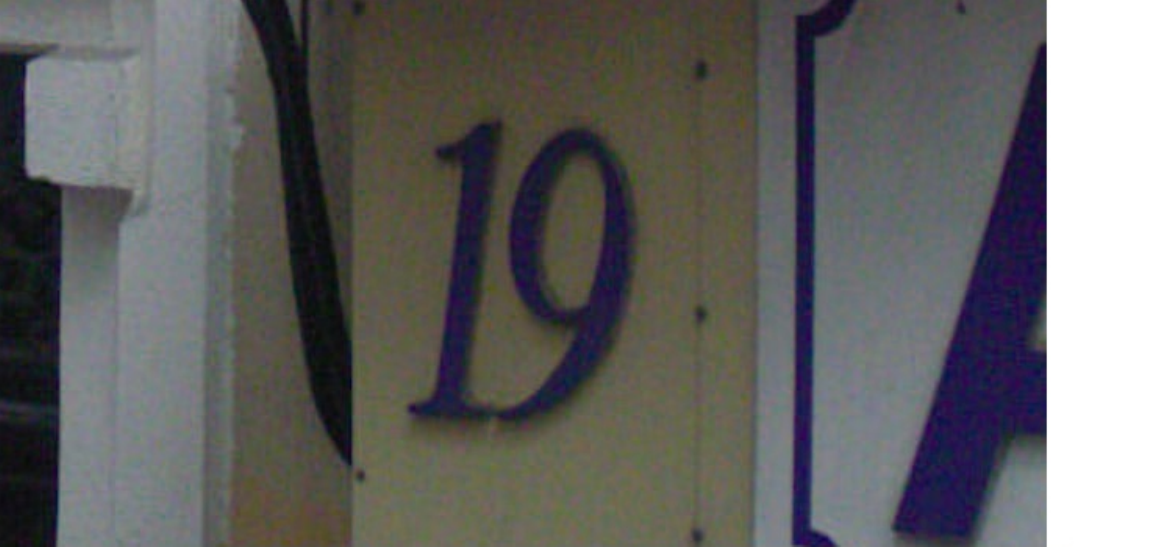}
        \subcaption{SVHN}
        \label{fig:svhn}
    \end{minipage}
    \caption{Sample images from the eight benchmark datasets used in our experiments.}
    \label{fig:dataset}
\end{figure}

\paragraph{Target Datasets.}
We evaluate classification accuracy on eight diverse datasets: Cars~\cite{krause20133d}, DTD~\cite{cimpoi2014describing}, EuroSAT~\cite{helber2019eurosat}, GTSRB~\cite{stallkamp2011german}, MNIST~\cite{lecun1998gradient}, RESISC45~\cite{cheng2017remote}, SUN397~\cite{xiao2010sun}, and SVHN~\cite{netzer2011reading}, as shown in Fig.~\ref{fig:dataset}. These datasets are commonly used in prior work on task arithmetic and model editing~\cite{ilharco2023editing, ortizjimenez2023tangent}, and provide a diverse benchmark for assessing cross-domain transferability.

\subsection{Task Vector Construction and Multi-Task Transfer}
For each target dataset \( \mathcal{D}_i \), we fine-tune the source model \( f^{(S)} \) to obtain a task-specific model \( f^{(S)}_i \), and define the task vector as the difference in parameters:
\[
\Delta \mathbf{W}_i = \mathbf{W}^{(S)}_i - \mathbf{W}^{(S)}, \quad i = 1, \dots, 8.
\]
All fine-tuning procedures use the AdamW optimizer~\cite{loshchilov2019adamw} with a learning rate of \(1.0 \times 10^{-5}\), weight decay of 0.1, and gradient clipping with a maximum norm of 1. We adopt cosine annealing as the learning rate scheduler.

Once task vectors are computed, we apply them cumulatively to the target model \( f^{(T)} \) using layer-wise orthogonal similarity transformations. Let each parameter matrix \( \mathbf{W} \) consist of \( L \) layers. For a specific layer \( l \), we denote the submatrix of the \( i \)-th task vector as \( \Delta \mathbf{W}_{i,l} \in \mathbb{R}^{d_l \times d_l} \). The transformed update for layer \( l \) is given by:
\[
\tilde{\mathbf{W}}^{(T)}_l = \mathbf{W}^{(T)}_l + \sum_{i=1}^8 \mathbf{U}_{i,l}^\top \Delta \mathbf{W}_{i,l}^{(S)} \mathbf{U}_{i,l},
\]
where \( \mathbf{W}^{(T)}_l \) is the target model's original parameter at layer \( l \), and \( \mathbf{U}_{i,l} \in \mathbb{O}(d_l) \) is a learned orthogonal matrix for task \( i \) and layer \( l \).

To train each orthogonal matrix \( \mathbf{U}_{i,l} \), we sample 100 images per target dataset and apply 10× data augmentation (horizontal flip, grayscale, blur), yielding 1,000 training examples. Training uses 100 mini-batches, with images refreshed at each batch.
Optimization settings follow those used for task vector construction: AdamW optimizer, learning rate \( 1.0 \times 10^{-5} \), weight decay 0.1, and cosine annealing.
This procedure is repeated for all layers \( l = 1, \dots, L \), producing a target model \( \tilde{f}^{(T)} \) that incorporates all adapted task vectors. We evaluate \( \tilde{f}^{(T)} \) on each of the eight tasks \( \mathcal{D}_1, \dots, \mathcal{D}_8 \).

\begin{table}[t]
\centering
\caption{Cross-model transfer accuracy (\%) on 8 target datasets. We compare Linear and LoRA task vectors with and without our orthogonal transformation. FS-Merge~\cite{kinderman2024foldable} and the unmodified target model ("Target only") are included as baselines.}
\label{tab:main_results}
\resizebox{\textwidth}{!}{
\begin{tabular}{lccccccccc}
\toprule
Method & Cars & DTD & EuroSAT & GTSRB & MNIST & RESISC45 & SUN397 & SVHN & Avg. \\
\midrule
Target model (only) & 74.54 & 52.29 & 47.37 & 40.78 & 50.71 &56.73 &64.17 & 27.90 & 51.81 \\
Target model (w/ FT) & 63.33 & 55.85 & 90.74 & 76.53 & 94.10 & 65.00 & 58.24 &79.24 &72.88\\
\midrule
Embed (w/o Ours) & {74.58} & 52.13 & 46.89 & 40.70 & 49.92 & 56.57 & {64.15} & 26.92 & 51.48\\
Embed (w/ Ours)  & 74.51 & {56.70} & {87.04} & 65.02 & {91.02} & {67.14} & 62.96 & 70.78 & {71.90}\\
\midrule
LoRA (w/o Ours) & 74.00 & 52.07 & 47.30 & 41.42 & 51.74 & 56.52 & 63.91 & 27.98 & 51.87\\
LoRA (w/ Ours)  & 66.66 & 51.49 & 85.93 & {70.51} & 90.06 & 60.89 & 59.80 & {73.86} & 69.90\\
\bottomrule
\end{tabular}
}
\end{table}

\subsection{Results}

Table~\ref{tab:main_results} shows the classification accuracy on the eight target datasets. As a baseline, we report the performance of the target model without task-specific adaptation (Target model only). This model reaches an average accuracy of 51.81\%.

We first examine task vectors based on the embedding layers of the source model (Embed). When we apply these vectors directly to the target model without transformation (w/o Ours), the average accuracy stays close to the baseline at 51.48\%. This result suggests that direct task vector transfer does not improve performance. In contrast, our orthogonal similarity transformation (w/ Ours) raises the average accuracy to 71.90\%. The method performs especially well on EuroSAT (87.04\%), MNIST (91.02\%), and SVHN (70.78\%).

We observe a similar trend with LoRA-based task vectors. Without transformation, the average accuracy is 51.87\%. After applying our transformation, the performance rises to 69.90\%. The improvements appear clearly on GTSRB (70.51\%) and SVHN (73.86\%).

These results show that transferring task vectors between models with different pre-training is not easy. Without proper alignment, the task vectors do not lead to gains in performance. Our method resolves this issue by using orthogonal similarity transformations to align task vectors with the target model’s parameter space. This alignment brings consistent and performance improvements across all datasets.

We also include the performance of the Target model (w/ FT), which uses the same few-shot data that we use for learning the orthogonal transformations. This model undergoes multi-task training across all eight target datasets and achieves the best average accuracy of 72.88\%. However, this approach ties all tasks together and removes modularity. In contrast, our method reaches nearly the same accuracy (71.90\% with Embed, 69.90\% with LoRA), while keeping task vectors modular and reusable. This feature represents a key strength of our approach—high transfer accuracy with minimal overhead.

We additionally test how two hyperparameters affect performance: the regularization coefficient $\alpha$ in the orthogonality loss and the scaling factor $\lambda$ for task vectors. Appendix~\ref{appendix:alpha} and Appendix~\ref{appendix:lambda} present these results. The accuracy improves most when both $\alpha$ and $\lambda$ take moderate values.

\section{Conclusion}

We have presented a method for transferring task vectors between models with different pre-training, a setting where conventional task arithmetic fails due to misaligned parameter spaces. Our approach introduces orthogonal similarity transformations to align task vectors from a source model to the coordinate system of a target model. This alignment preserves both the norm and rank of task vectors and enables effective reuse of task-specific updates.

Through experiments on the eight diverse target datasets, we demonstrate that our method substantially improves cross-model transfer accuracy compared to direct task vector application. In particular, task vectors constructed from embedding or LoRA modules achieve strong performance when transformed using our method, nearly matching that of full-model few-shot fine-tuning. Importantly, our approach preserves the modularity and compositionality of task vectors, enabling efficient application across tasks without retraining the full model.

These results highlight the effectiveness of our alignment-based approach to task arithmetic, and suggest that orthogonal transformations can serve as a simple yet powerful tool for bridging parameter-space mismatches in pre-trained neural networks. As future work, we plan to extend our method to a zero-shot setting, where alignment is performed without access to any labeled examples from the target domain. This would further enhance the practicality of task vector transfer in real-world scenarios.

\bibliographystyle{unsrt}  
\bibliography{references}

\appendix

\section{Appendix: Orthogonal Transformation Using Full parameters}\label{appendix:full weight}

We investigate whether task vectors constructed from all parameters of the pre-trained model (i.e., full fine-tuning) can benefit from our orthogonal similarity transformation.

In this setting, instead of restricting task vector computation to embedding or LoRA parameters, we fine-tune all parameters of the source model using few-shot data from each task to compute a full-parameter task vector. We then align each layer of this task vector to the target model via orthogonal similarity transformations, following the same procedure as in our proposed method.

\begin{table}[h]
\centering
\caption{Transfer accuracy (\%) using task vectors constructed from full-model fine-tuning followed by orthogonal transformation.}
\label{tab:full_weight_alignment}
\begin{tabular}{ccccccccc}
\toprule
Cars & DTD & EuroSAT & GTSRB & MNIST & RESISC45 & SUN397 & SVHN & Avg \\
\midrule
31.65 & 36.60 & 64.70 & 32.76 & 51.73 & 40.97 & 37.33 & 36.80 & 41.57 \\
\bottomrule
\end{tabular}
\end{table}

As shown in Table~\ref{tab:full_weight_alignment}, this full-parameter setting performs substantially worse than our selective methods (e.g., 71.90\% average with Embed). We interpret this degradation as follows.

Full-model fine-tuning introduces high-dimensional, noisy task vectors that may encode spurious or overfit patterns, making alignment less effective. Furthermore, aligning all parameter matrices increases the number of parameters involved in transformation, which may lead to overfitting or optimization instability in few-shot settings. In contrast, task vectors computed from specific parameter subsets (such as attention embeddings or LoRA layers) offer compact and more informative directions in parameter space, which are easier to align and transfer. 

\section{Ablation on the Regularization Coefficient \texorpdfstring{$\alpha$}{lambda}}
\label{appendix:alpha}

We conduct an ablation study to investigate the effect of the regularization coefficient $\alpha$ in our objective function:
\[
\mathcal{L}_{\mathrm{total}}
= \sum_{l=1}^L 
   \mathcal{L}_{\mathrm{CE}}\Bigl(\mathbf{W}^{(T)}_l + \mathbf{U}_l^\top \Delta \mathbf{W}^{(S)}_l\,\mathbf{U}_l\Bigr)
\;+\;\alpha
\sum_{l=1}^L 
   \bigl\|\mathbf{U}_l^\top\mathbf{U}_l - \mathbf{I}\bigr\|_F^2.
\]

Since Embed-based task vectors slightly outperform LoRA-based ones in our main experiments (Table~\ref{tab:main_results}), we focus this ablation study on the Embed setting only. All other experimental settings follow those described in the main text.

\begin{table}[h]
\centering
\caption{Ablation study on the regularization coefficient $\alpha$ using Embed-based task vectors.}
\label{tab:lambda_ablation}
\begin{tabular}{c|ccccccccc}
\toprule
$\alpha$ & Cars & DTD & EuroSAT & GTSRB & MNIST & RESISC45 & SUN397 & SVHN & Avg. \\
\midrule
0.3 & 74.29 & 56.28 & 87.41 & 69.97 & 88.04 & 67.87 & 61.83 & 70.40 & 72.01 \\
0.8 & 73.03 & 55.00 & 86.93 & 68.09 & 88.20 & 67.46 & 63.08 & 69.93 & 71.46 \\
1.0 & 74.51 & 56.70 & 87.04 & 65.02 & 91.02 & 67.14 & 62.96 & 70.78 & 71.90 \\
2.0 & 72.59 & 54.63 & 84.96 & 68.22 & 88.36 & 65.95 & 63.39 & 72.45 & 71.32 \\
\bottomrule
\end{tabular}
\end{table}

The results in Table~\ref{tab:lambda_ablation} show that performance is generally robust to changes in $\alpha$ within a reasonable range. The best average accuracy is observed at $\alpha = 0.3$, while $\alpha = 1.0$—used in the main experiments—also performs competitively. Larger values may overly constrain the orthogonality, while smaller values may insufficiently regularize the transformation.

\section{Ablation on Task Vector Scale \texorpdfstring{$\lambda$}{lambda}}
\label{appendix:lambda}

We investigate the effect of scaling the task vector magnitude using a scalar coefficient $\lambda$. Specifically, we modify the task vector as follows:
\[
\Delta \mathbf{W}_i = \lambda(\mathbf{W}^{(S)}_i - \mathbf{W}^{(S)}), \quad i = 1, \dots, 8,
\]
where $\Delta \mathbf{W}_i$ denotes the task vector for the $i$-th dataset, and $\mathbf{W}^{(S)}_i$ and $\mathbf{W}^{(S)}$ are the parameters of the fine-tuned and pre-trained source models, respectively.

This experiment is conducted under the same conditions as the main experiments. We fix the orthogonal transformation scale factor $\alpha = 1.0$, and evaluate only the Embed configuration, which outperformed LoRA in the main results.

Table~\ref{tab:ablation_lambda} shows the classification accuracy across eight datasets with different $\lambda$ values. We observe that the default setting $\lambda = 1.0$ yields the highest average performance (71.90\%). Smaller values such as $\lambda = 0.3$ underperform due to insufficient adaptation, while overly large values such as $\lambda = 2.0$ may lead to overfitting or misalignment in the transformed parameter space. These results suggest that the scale of task vectors plays a critical role in successful transfer.

\begin{table}[h]
\centering
\caption{Effect of task vector scaling parameter $\lambda$ on transfer accuracy (\%). Evaluation uses Embed-based vectors with $\alpha=1.0$.}
\label{tab:ablation_lambda}
\begin{tabular}{c|ccccccccc}
\toprule
$\lambda$ & Cars & DTD & EuroSAT & GTSRB & MNIST & RESISC45 & SUN397 & SVHN & Avg. \\
\midrule
0.3 & 73.30 & 54.31 & 80.56 & 62.78 & 80.86 & 62.84 & 64.33 & 65.93 & 68.11 \\
0.8 & 72.93 & 53.83 & 87.33 & 68.31 & 86.04 & 64.60 & 63.06 & 67.16 & 70.41 \\
1.0 & 74.51 & 56.70 & 87.04 & 65.02 & 91.02 & 67.14 & 62.96 & 70.78 & {71.90} \\
2.0 & 70.55 & 52.77 & 89.30 & 70.32 & 88.13 & 64.13 & 61.09 & 72.43 & 71.09 \\
\bottomrule
\end{tabular}
\end{table}

\end{document}